\documentclass[11pt]{article}

\usepackage[preprint]{acl}

\usepackage{times}
\usepackage{latexsym}

\usepackage[T1]{fontenc}

\usepackage[utf8]{inputenc}

\usepackage{microtype}

\usepackage{inconsolata}

\usepackage{placeins}

\usepackage{graphicx}
\usepackage{amsmath}
\usepackage{multirow}
\usepackage{subcaption}
\usepackage{enumitem}
\usepackage{booktabs}
\usepackage{placeins}
\usepackage{needspace}
\usepackage{amsthm}
\newtheorem*{example}{Example}
\providecommand{\tightlist}{%
  \setlength{\itemsep}{0pt}\setlength{\parskip}{0pt}}
\title{Logic-Parametric Neuro-Symbolic NLI: \\Controlling Logical Formalisms for Verifiable LLM Reasoning}

\author{Ali Farjami$^{1}$, Luca Redondi$^{1,2}$, Marco Valentino$^{3}$\\ 
$^{1}$University of Luxemburg\quad
$^{2}$Ruhr-Universtät Bochum\quad
$^{3}$University of Sheffield\\
  \texttt{ali.farjami@uni.lu}\quad
  \texttt{luca.redondi@ruhr-uni-bochum.de}\\ 
\texttt{m.valentino@sheffield.ac.uk}\\
}

\begin{document}
\maketitle
\begin{abstract}
Large language models (LLMs) and theorem provers (TPs) can be effectively combined for verifiable natural language inference (NLI). However, existing approaches rely on a fixed logical formalism, a feature that limits robustness and adaptability.
We propose a logic-parametric framework for neuro-symbolic NLI that treats the underlying logic not as a static background, but as a controllable component. Using the LogiKEy methodology, we embed a range of classical and non-classical formalisms into higher-order logic (HOL), enabling a systematic comparison of inference quality, explanation refinement, and proof behavior.
We focus on normative reasoning, where the choice of logic has significant implications. In particular, we compare logic-external approaches, where normative requirements are encoded via axioms, with logic-internal approaches, where normative patterns emerge from the logic’s built-in structure. Extensive experiments demonstrate that logic-internal strategies can consistently improve performance and produce more efficient hybrid proofs for NLI.
In addition, we show that the effectiveness of a logic is domain-dependent, with first-order logic favouring commonsense reasoning, while deontic and modal logics excel in ethical domains. Our results highlight the value of making logic a first-class, parametric element in neuro-symbolic architectures for more robust, modular, and adaptable reasoning.
\end{abstract}

\section{Introduction}
 Large Language Models (LLMs) have made impressive strides in natural language inference (NLI), enabling plausible and fluent explanations across a wide range of tasks \cite{liu2025logical,cheng2025empowering}. Yet when it comes to inference that must be logically valid, generalizable, and trustworthy, such as in legal, ethical, or regulatory contexts, existing LLM systems often fall short \cite{li2023dark,hadi2023large,bender2021dangers}. 
 
 A potential solution is provided by neuro-symbolic architectures \cite{bhuyan2024neuro,garcez2023neurosymbolic}, where LLMs are combined with external theorem provers (TPs) for formal verification and refinement \cite{quan-etal-2025-faithful,quan-etal-2025-peirce,quan-etal-2024-verification,pan-etal-2023-logic,olausson-etal-2023-linc}. However, a key limitation of existing approaches lies in how logic is handled: most neuro-symbolic systems fix a single logic, typically first-order logic (FOL), treating it as a static background layer, rather than an adaptable component.

\begin{figure*}[t]
    \centering
    \includegraphics[width=\textwidth]{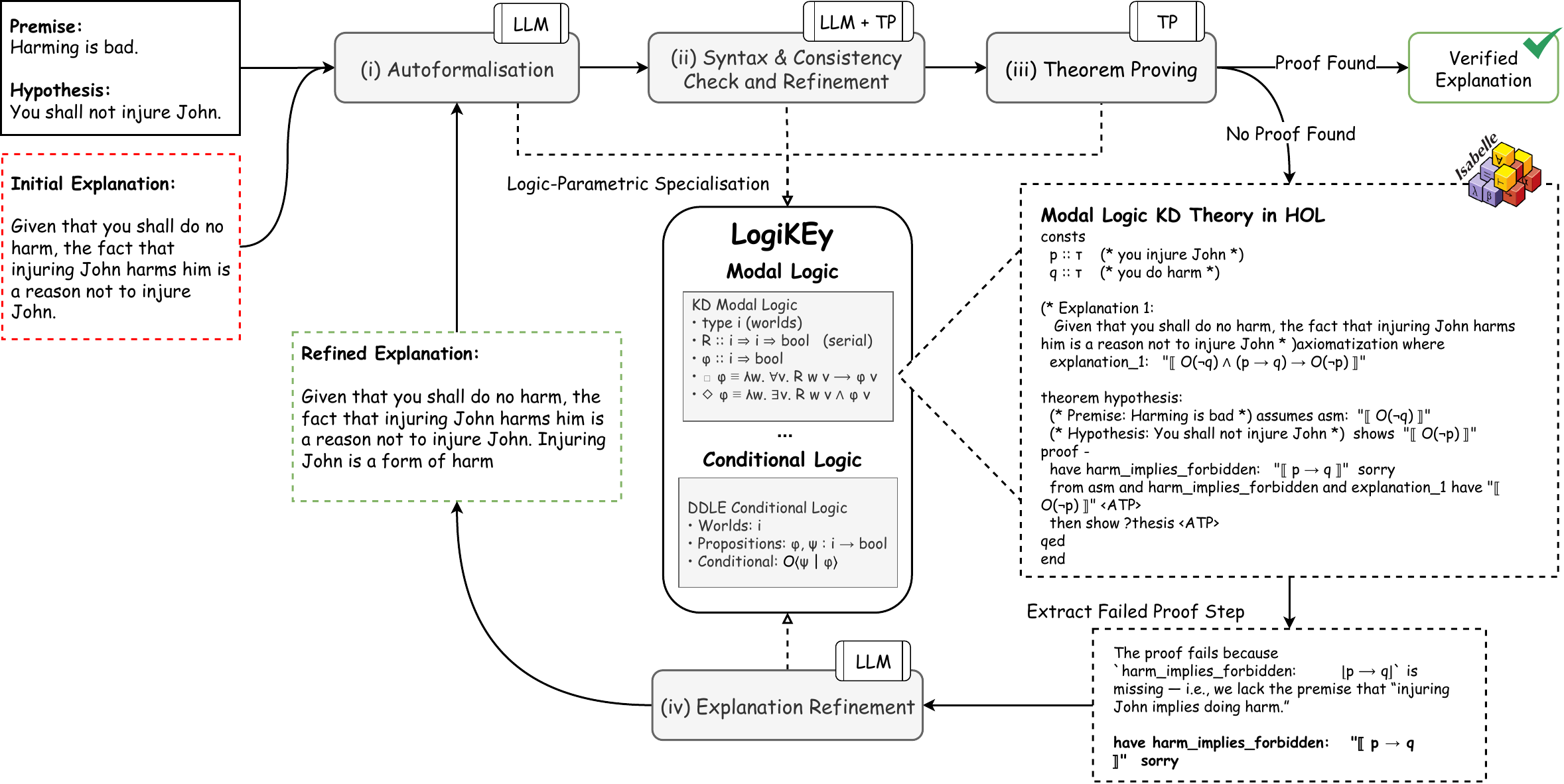}
    \caption{Illustration of the logic-parametric neuro-symbolic NLI framework with LLMs. The framework generalizes neuro-symbolic architectures via \textit{LogiKEy}, embedding classical and non-classical logics into higher-order logic (HOL). This enables the integration of LLMs and theorem provers (TPs) using diverse logical formalisms for iterative explanation refinement across tasks and domains.
    }
    \label{fig:logikey}
\end{figure*}

In this paper, we challenge this assumption. We argue that \textit{logic itself should be treated as a parameter} in neuro-symbolic reasoning. Different logical systems afford different reasoning patterns: modal logics can natively express obligation and permission, conditional logics handle exceptions and context shifts, and first-order event logics excel at encoding specific instances. By enabling \textit{logic-parametric architectures}, we can systematically explore how the structure of a logic affects LLM-driven reasoning, and when one logic may outperform another.

To ground this inquiry, we focus on \textit{normative reasoning} as a domain where the choice of logic is especially impactful~\cite{gabbay13}. In contexts like ethics, law, and policy, reasoning involves obligations, permissions, prohibitions, exceptions, and violations. These concepts are often difficult to capture using traditional FOL. For example, the inference from
\textit{``you are obliged to submit the assignment''} to
\textit{``you are permitted to submit the assignment''}, while intuitive, cannot be derived from FOL alone without adding external axioms. In contrast, the modal logic KD includes this as a built-in axiom ($O\varphi \rightarrow P\varphi$)~\cite{chellas1980modal}.
This leads to a central distinction between \textit{logic-external reasoning}, where normative rules are added explicitly as axioms in the domain theory (e.g., in FOL); and
\textit{logic-internal reasoning}, where rules are embedded in the logic itself as structural principles (e.g., modal axioms, conditional operators).

We build on the \textit{LogiKEy} methodology~\cite{J48}, which supports semantic embeddings of classical and non-classical logics in higher-order logic (HOL). This infrastructure enables us to integrate various logics into a hybrid LLM-TP pipeline and compare their behaviour in formalising and verifying explanations for NLI.

Our investigation centers on the following research questions:
(\textit{RQ1}) How does the choice of logical formalisms affect LLM-driven theorem proving for NLI? (\textit{RQ2}) How do logic-internal reasoning strategies compare to logic-external approaches? (\textit{RQ3}) Can logic-parametric architectures improve proof economy, explanation refinement, and verification?



To systematically investigate these questions, we first introduce a new dataset for deontic explanation in NLI, called \textit{B}ioethical \textit{E}xplanations and \textit{N}ormative \textit{R}easoning (\textit{BENR})\footnote{Code and dataset will be released upon publication.}
, designed to expose the internal structure of normative explanations \cite{Goble13}. 
Using \textit{BENR}, we conduct a logic-parametric evaluation by extending and generalizing established LLM-driven neuro-symbolic frameworks for the verification and refinement of NLI explanations \cite{quan-etal-2025-faithful,quan-etal-2024-verification}. 

Our results show that logic-internal approaches consistently outperform logic-external methods in normative reasoning tasks across different LLMs, including GPT-4o \cite{bubeck2023sparks} and DeepSeek-V1 \cite{bi2024deepseek}. In particular, using the modal logic KD, we can achieve the highest explanation refinement rate (up to 77.67\%), converging with fewer iterations, and substantially reducing inference cost compared to first-order logic (FOL). 
In contrast, FOL exhibits higher syntactic robustness but validates significantly fewer explanations, highlighting a trade-off between expressive adequacy and stability. Moreover, our results highlight that the effectiveness of a logic is domain-dependent, with FOL favoring commonsense reasoning while deontic and modal logics favoring ethical domains.

\begin{table*}[t]
\centering
\small
\renewcommand{\arraystretch}{1.3}
\begin{tabular}{p{0.12\textwidth}|p{0.2\textwidth}p{0.2\textwidth}p{0.2\textwidth}|p{0.15\textwidth}}
\toprule
\textbf{Pipeline Stage} & \textbf{FOL} & \textbf{Modal Logic} & \textbf{Conditional Logic} & \textbf{Purpose} \\
\midrule
\textbf{Syntactic Parsing} & Identify subject, verb, object & Identify modal keywords: \textit{must, may, ought} & Identify conditional clauses: \textit{if-then, unless} & Structure extraction \\
\midrule
\textbf{Formalization} & $\text{Agent}(e,x) \land \text{Verb}(e) \land \text{Patient}(e,y)$ & $\bigcirc(\text{verb})$ or $P(\text{verb})$ & $\bigcirc(\psi/\varphi)$~or $P(\psi/\varphi)$ & NL $\rightarrow$ logic mapping \\
\midrule
\textbf{Proof Sketch} & Use $\land$, $\rightarrow$, $\forall$, $\exists$ & Use $\bigcirc$, $P$,  $\Box$, $\Diamond$  & Use $\bigcirc(-/-)$, $P(-/-)$ & Guided proving \\
\midrule
\textbf{Refinement} & ``Missing premise about $\text{Agent}$'' & ``Modal Axiom not satisfied: $\bigcirc \varphi \rightarrow P\varphi$'' & ``Conditional norm not detachable: $\varphi \land \bigcirc(\psi/\varphi)$ '' & Feedback, error correction \\
\bottomrule
\end{tabular}
\caption{Logic-parametric adaptation across pipeline stages. Each stage of the pipeline is tailored to a specific logic, respecting its syntax and semantics.}
\label{tab:prompt-variants}
\end{table*}

To summarize, our contributions are as follows:
\begin{enumerate}
    \item We propose a \textit{logic-parametric neuro-symbolic framework} that treats the choice of logic as an explicit design dimension in LLM-driven theorem proving for NLI.
    \item We introduce \textit{BENR}, a new dataset for deontic explanation in NLI, designed to expose the internal structure of normative reasoning and explanation.
    \item We provide an extensive empirical analysis showing that logic-internal reasoning improves robustness, proof economy, and explanation refinement compared to logic-external strategies. Overall, we demonstrate that the choice of logic has a decisive impact on LLM-driven neuro-symbolic systems, establishing the foundations for a more effective, adaptable, and modular integration.
\end{enumerate}



\section{Logic-Parametric Explanation Verification for NLI}

Given an NLI problem consisting of a hypothesis $h$, a premise $p$, and an explanation $e$, each expressed in natural language, $e$ is defined as a logically valid explanation if $p \cup e \models h$. To verify this, the triple $\{p, e, h\}$ is mapped into a set of logical formulae. The formulae for $p$ and $e$ constitute the theory $\Theta$, while $h$ is mapped to a target formula $\psi$. A theorem prover can then be used to determine if $\Theta \models \psi$, thereby validating the explanation.

In this work, we extend and generalize the neuro-symbolic framework introduced by \citet{quan-etal-2025-faithful}, proposing a modular, logic-parametric pipeline designed to accommodate diverse formalisms, such as modal logic and conditional logic, in Isabelle/HOL . As illustrated in Figure \ref{fig:logikey}, the pipeline orchestrates the interaction between LLMs and theorem prover through four key stages:

\paragraph{(i) Autoformalization} 
An LLM maps the natural language triple $\{p, h, E\}$ into a set of formulae $\Phi$. This stage involves syntactic parsing followed by translation into a target formal language and the definition of a proof sketch that can be validated by a theorem prover \cite{quan-etal-2024-verification}. Unlike prior work restricted to first-order logic, our framework parameterizes this stage by the target logic $\mathcal{L}$. This results in a formal theory $\Theta_{\mathcal{L}}$ and a goal formula $\psi_{\mathcal{L}}$ representing the hypothesis.

\paragraph{(ii) Syntax \& Consistency Check} 
After the formalization, $\Theta_{\mathcal{L}}$ and $\psi_{\mathcal{L}}$ undergo an automated syntactic and consistency check. This ensures that the LLM-generated formalizations are syntactically valid for the Isabelle/HOL environment and that the premises are non-contradictory (i.e., $p \cup e \not\models \bot$), preventing vacuous entailment.

\paragraph{(iii) Theorem Proving} 
The formal theory is processed by the Isabelle/HOL interactive theorem prover~\cite{Isabelle}. Depending on the chosen logic module, the system utilizes specialized axiomatizations. The prover attempts to derive $\Theta_{\mathcal{L}} \vdash \psi_{\mathcal{L}}$ using automated theorem proving tools integrated in Isabelle.

\paragraph{(iv) Explanation Refinement} 
If the proof fails, the framework extracts the failed proof step returned by the theorem prover. This symbolic feedback identifies missing premises or logical gaps (e.g., a missing bridge rule $p \to q$ as shown in Figure \ref{fig:logikey}). The feedback is then provided to the LLM to generate a revised explanation $e'$ following a refinement strategy. This cycle iterates for $t$ steps or until the explanation can be successfully verified.

\subsection{Semantic Embeddings via LogiKEy}

Our implementation is based on the \textit{LogiKEy} methodology~\cite{J48}, which supports semantic embeddings of a wide variety of logical systems into higher-order logic (HOL). These embeddings preserve the semantics of each target logic within a unified formal meta-language, allowing them to share infrastructure such as theorem provers (e.g., Isabelle/HOL), model finders (e.g., Nitpick~\cite{Nitpick}), and proof assistants (e.g., Sledgehammer~\cite{blanchette2013extending}). Rather than translate individual formulas from one logic to another, LogiKEy treats each logic as a first-class module, with its own axioms, operators, and inference constraints, all embedded semantically within HOL. This enables logic-parametric experimentation within a uniform and verifiable framework.

\subsection{Supported Logics}

The LogiKEy framework supports a range of logics~\cite{J53} relevant to normative reasoning and beyond:
\paragraph{Modal Logic KD} Expresses effectively the logical relations between obligation, permission and prohibition and supports basic modal inference \cite{von1951deontic}. The language of {\bf K} is obtained by supplementing the language of propositional logic (PL) with a  modal operator $ \bigcirc$. It is generated as follows:
       $$ \varphi ::= p|\neg \varphi| \varphi \vee \varphi | \bigcirc\varphi  $$
     $P$ is the dual of $\bigcirc$, viz. $ P \varphi =_{df} \neg \bigcirc \neg\varphi$.  Modal Logic {\bf KD} is the extension of modal logic  {\bf K} with the axiom \textbf{D}: $ \bigcirc  \varphi \rightarrow P \varphi $ that  captures the intuition that obligations imply permissions.

\paragraph{Conditional Logic DDLE} Substitutes the standard possible-worlds semantics used by {\bf KD} with a preference-based semantics~\cite{aaqvist1984deontic,parent2021preference}. In the possible world semantics, we specify the acceptable world accessible from each world, and define obligation accordingly. On the other hand, in preference-based semantics all words are ordered from the ideal world to the (morally) worst one. This enables to express contrary-to-duty obligations~\cite{chisholm63:_contr_duty_imper_deont_logic}, i.e.: obligation that becomes compelling in sub-optimal worlds because some other obligation has been violated. The language of {\bf DDLE} is obtained by adding the following operators to the language of propositional logic: $\Box$ (for necessity); $\Diamond$ (for possibility); and $\bigcirc (-/-)$ (for conditional obligation) ; $P (-/-)$ (for conditional permission).  $\bigcirc (\psi/\varphi)$ is read ``If $\varphi$, then $\psi$ is obligatory'', and  $P(\psi/\varphi)$ is read ``If $\varphi$, then $\psi$ is permitted.'' 
    
    


\paragraph{Conditional Logic with Factual Detachment} Violating obligations does not make them vanish. Therefore, in contrary-to-duty scenario, two different meaning of ``ought'' emerge: on one side we have what ideally should be the case, on the other, what actually should be the case, given that a violation already occurred. Carmo and Jones Dyadic Deontic Logic~\cite{CJ13}, we shall refer to it as {\bf DDL\_CJ}, is capable of representing both without ambiguities. The set of {\bf DDL\_CJ} formulas extends the set of conditional logic formulas (as discussed in system {\bf DDLE}) with the following:
\begin{itemize}
\tightlist
    \item $\Box \varphi$ --- \textit{in all worlds}
    \item $\Box_{a} \varphi$ --- \textit{in all actual versions of the current world}
    \item $\Box_{p} \varphi$ --- \textit{in all potential versions of the current world}
    \item $\bigcirc_{a} \varphi$ --- \textit{monadic deontic operator for actual obligation}
    \item $\bigcirc_{p} \varphi$ --- \textit{monadic deontic operator for primary obligation}
\end{itemize}

By embedding these logics in HOL~\cite{C71, J45, fth20}, we unify them within a common reasoning framework while preserving their distinct inferential properties.

\section{Empirical Setup}
\subsection{The BENR Dataset}


To test the model's capability to produce valid deontic explanations, we construct a dataset called \textit{B}ioethical \textit{E}xplanations and \textit{N}ormative \textit{R}easoning (\textit{BENR}). The focus of BENR is to explore the different reasoning patterns at work in ethical reasoning. Compared to the existing alternatives, it displays one main distinctive feature. 
Datasets often aim for simplicity: cases are described at a high level of abstraction and contextualized within specific scenarios \cite{hendrycks2021aligning, forbes2020social}. In contrast, 
we are not interested in the scenarios to which ethical reasoning is applied, but rather in the structure of the reasoning itself,\footnote{\cite{emelin2021moral} for instance, presents rich scenarios, but simple ethical reasoning.} including the different composing patterns 
, and the way they are combined when a moral evaluation is performed.  


To achieve this goal, we target a distinctively complex subfield of applied ethics: \textit{(Bio)ethics}. The dataset includes a total of 103 examples. A good part of the dataset (47 cases) includes reasoning patterns that are typical of (bio)ethical reasoning, such as the instantiation of prima-facie reasons from general principles and the resolution of conflict between them~\cite{Goble13}. The other 56 cases include reasoning patterns that, although relevant in ethical reasoning, are not peculiar to ethics. These include epistemic default reasoning and reasoning about deontic modalities. 
 For the second subset, the scope of our dataset overlaps with other datasets. Therefore, we build upon previous resources, adapting existing examples to our format. In particular, we adapt classical logic problems and problems about modalities from \cite{holliday2024conditional}, and commonsense and default reasoning problems from e-SNLI \cite{camburu2018snli}. 
Overall, the cases in our dataset exhibit the following format:

\begin{example}[Autonomy requires competent choice]
\label{ex:autonomy}
\textbf{Premise.} A patient refuses a simple and life-saving treatment. The patient is severely confused because of a high fever. You should respect others’ autonomy. Promoting others’ wellbeing is good.
\medskip

\textbf{Hypothesis.}
You ought to give this treatment.
\medskip

\textbf{Explanation.}

\begin{enumerate}
\tightlist
    \item Given that you should respect autonomy, the patient’s refusal would normally be a reason not to treat.
    \item Given that promoting others’ wellbeing is good, the life-saving benefit is a reason to treat.
    \item But the refusal was made without mental competence, so it does not express an autonomous decision.
    \item Thus the reason not to treat is undercut, and the reason to treat remains.
\end{enumerate}
\end{example}




The explanation indicates the reasoning steps that bridge from the premises to the hypothesis. In the present example, these include the detachment of prima-facie reasons from general principles, and the resolution of conflict between prima-facie reasons by means of an undercut. 

\subsection{Models}

To support logic-parametric theorem proving, we extend Faithful-Refiner \cite{quan-etal-2025-faithful}, an explanation refinement framework for NLI originally designed for Neo-Davidsonian first-order logic (FOL) formalization in Isabelle/HOL. While their prompts effectively guide LLMs toward event-based representations, they assume a fixed logical substrate. Our extension generalises the framework to accommodate a range of classical and non-classical logics embedded via the LogiKEY framework. Key differences across logical formalisms are summarised in Table \ref{tab:prompt-variants}.

We evaluate two state-of-the-art LLMs: \textit{GPT-4o} \cite{bubeck2023sparks} and \textit{DeepSeek-V1} \cite{bi2024deepseek} . Both models are used in their default configurations with identical prompts to ensure fair comparison of their reasoning capabilities across logical frameworks. We evaluate the models using up to $t=3$ refinement iterations.

\subsection{Logical Formalisms}

We evaluate the framework across four logical formalisms -- \textit{FOL}, \textit{KD}, \textit{DDLE}, and \textit{DDL\_CJ} -- over the diverse set of reasoning tasks and domains in BENR, including classical logic, commonsense reasoning, default reasoning, modalities, and bioethical reasoning. 

\begin{figure}[t]
    \centering
    \includegraphics[width=\linewidth]{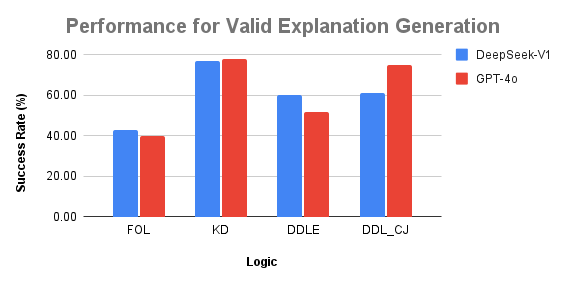}
    \caption{Success rates for valid explanation generation. }
    \label{fig:success-rates}
\end{figure}

\begin{figure}[t]
    \centering
    \includegraphics[width=\linewidth]{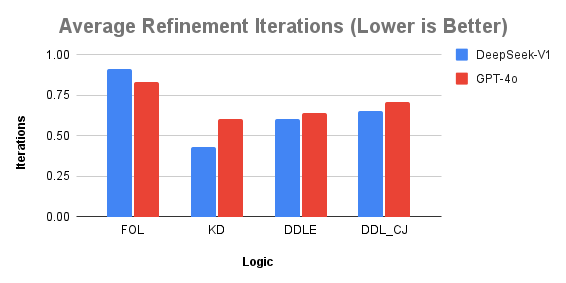}
    \caption{Average number of refinement iterations required to reach a valid explanation.}
    \label{fig:iterations}
\end{figure}

\begin{figure*}[t]
    \centering
    \includegraphics[width=\textwidth]{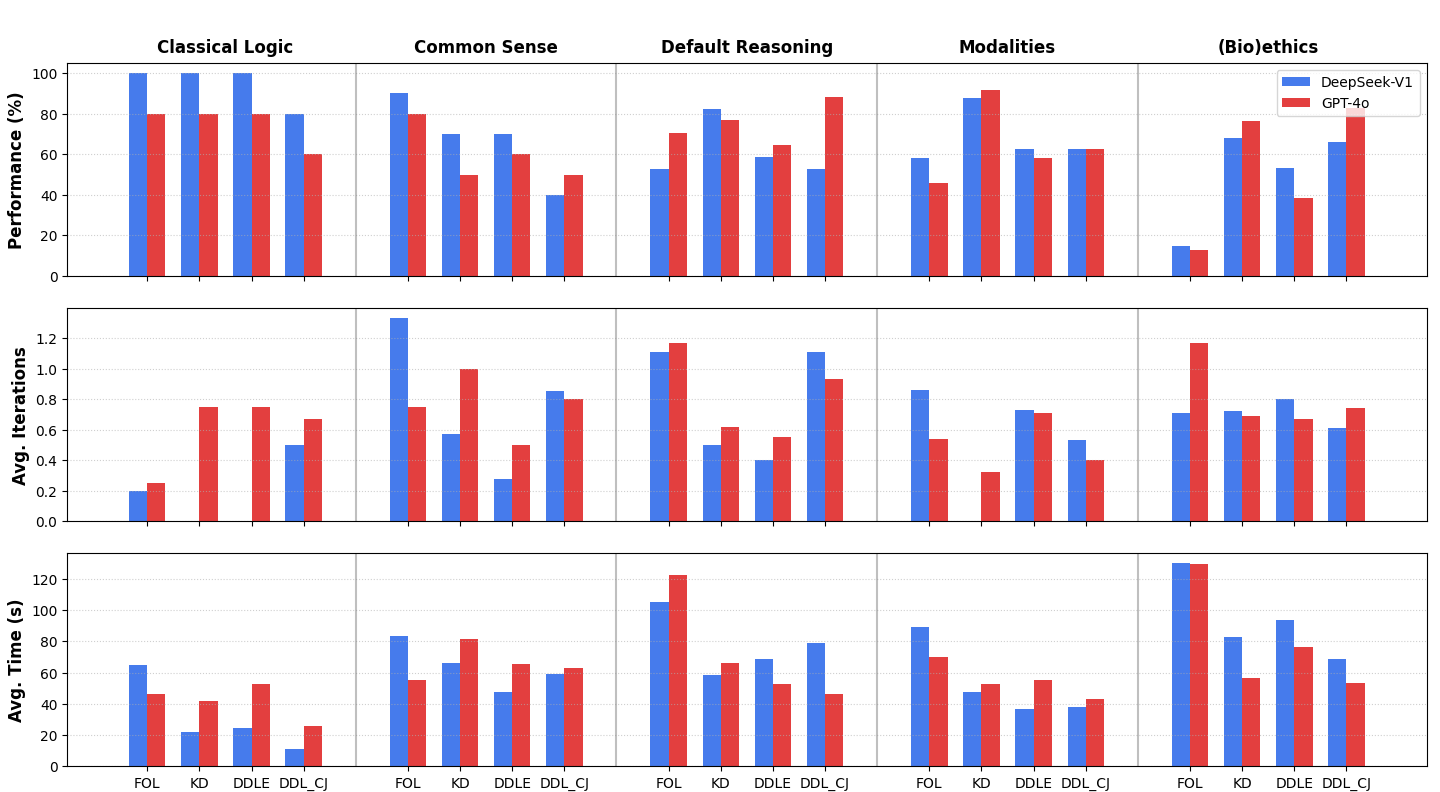}
    \caption{Explanation refinement performance across logical frameworks and domains for both DeepSeek-V1 and GPT-4o. Solving time is averaged over successful runs only.}
    \label{fig:domain-performance-percent-capped}  
\end{figure*}

\begin{figure}[t]
    \centering
    \includegraphics[width=\linewidth]{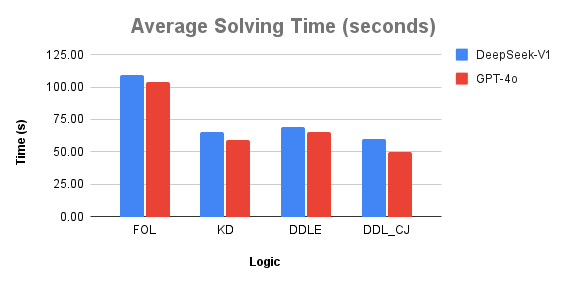}
    \caption{Average solving time (seconds) over successful explanation refinements.}
    \label{fig:runtime}  
\end{figure}

\begin{figure}[t]
    \centering
    \includegraphics[width=\linewidth]{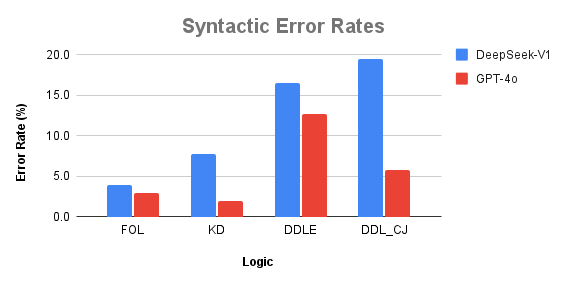}
    \caption{Syntactic error rates.}
    \label{fig:syntactic-errors}  
\end{figure}

\section{Results}

Performance is analysed along four complementary dimensions: overall explanation success rate (Figure~\ref{fig:success-rates}), refinement efficiency (Figure~\ref{fig:iterations}), computational efficiency (Figure~\ref{fig:runtime}), robustness to syntactic failure (Figure~\ref{fig:syntactic-errors}), and domain-specific behaviour (Figure~\ref{fig:domain-performance-percent-capped}).

\paragraph{Explanation Success Rates}

Figure~\ref{fig:success-rates} presents the success rates of both DeepSeek and GPT-4o across the four logical frameworks. The success rate represents the percentage of test cases for which each model-logic combination successfully produced a valid explanation for the NLI problem. 
We found that GPT-4o with KD achieves the highest success rate (77.67\%), while both models struggle with {\it FOL}.


\paragraph{Refinement Efficiency}

Refinement efficiency reflects how quickly a logic converges to a valid explanation. Figure~\ref{fig:iterations} focuses exclusively on refinement depth.
{\it KD} consistently reaches valid explanations with fewer refinement steps than the other logics. 

\paragraph{Computational Efficiency}

Figure~\ref{fig:runtime} reports average solving time per logic.
{\it FOL} exhibits the highest computational cost, reflecting the expense of first-order reasoning combined with refinement. {\it KD} significantly reduces runtime, while {\it DDL\_CJ} achieves the lowest average solving time despite operating in a more expressive deontic framework. Notably, refinement depth and runtime are not perfectly correlated: expressive logics may require additional refinement steps but still achieve competitive solving times once syntactically valid explanations are obtained.

\paragraph{Robustness to Syntactic Failure}

In addition to successful refinements, syntactic errors can constitute a major failure mode. Figure~\ref{fig:syntactic-errors} summarises the observed robustness of each logic.
Deontic logics, particularly {\it DDLE} and {\it DDL\_CJ}, are more sensitive to malformed intermediate representations, especially in default reasoning, modal reasoning, and bioethical scenarios. {\it FOL} shows higher syntactic robustness but validates fewer explanations overall, highlighting a trade-off between stability and expressive coverage.

\begin{table*}[t]
\centering
\tiny
\begin{tabular}{p{0.475\textwidth}|p{0.475\textwidth}}
\toprule
\textbf{Normative/Deontic Focus} & \textbf{Descriptive/Factual Focus} \\
\midrule
\textbf{DDL\_CJ}: 
\textit{Refinement Strategy:} ``The failed proof step indicates that the logical chain from the refusal not expressing an autonomous decision to the obligation to give treatment is incomplete. We need to refine the explanatory sentences to establish a clear link between the lack of mental competence, the undercutting of the refusal as a reason not to treat, and the obligation to provide treatment.''
\newline\newline
\textbf{KD}: 
\textit{Refinement Strategy:} ``The failed proof step indicates that the logical connection between the life-saving benefit being a reason to treat and the obligation to give the treatment is missing or incomplete. We need to refine the explanatory sentences to establish a clear link between the life-saving benefit, the undercutting of the refusal, and the obligation to give the treatment.''
&
\textbf{FOL}: 
\textit{Refinement Strategy:} ``The failed proof step indicates a logical error in deriving 'the life-saving benefit is a reason to treat' from the given explanations. The issue seems to stem from the lack of a direct connection between the patient's mental competence and the life-saving benefit as a reason to treat. We need to refine the explanations to establish a clearer connection between the patient's mental state and the justification for treatment.''
\newline\newline
\textbf{DDLE}: 
\textit{Refinement Strategy:} ``The failed proof step indicates that the assumption (the patient is not competent) could not be established. The logical connection between the patient's mental competence and the refusal not expressing an autonomous decision is not sufficiently supported. We need to refine the explanatory sentences to make this connection explicit and ensure that the logical chain from the patient's mental competence to the treatment decision is clear and complete.''
\\
\bottomrule
\end{tabular}

\begin{tabular}{p{0.475\textwidth}|p{0.475\textwidth}}
\toprule
\textbf{Normative Refinement} & \textbf{Descriptive Refinement} \\
\midrule
 \textbf{DDL\_CJ}: ``When a refusal is not autonomous, the obligation to promote others' wellbeing by providing life-saving treatment takes precedence.''
\newline\newline
\textbf{KD}: ``When the life-saving benefit is the primary consideration and the reason not to treat is undercut, you ought to give the treatment.''

&
\textbf{FOL}: ``Given that you should respect autonomy, the patient's refusal would normally be a reason not to treat, \emph{unless the refusal is made without mental competence}.''
\newline\newline
\textbf{DDLE}: ``A refusal that does not express an autonomous decision is not a valid reason not to treat.''
\\
\bottomrule
\end{tabular}
\caption{(Top) Refinement strategies and resulting statements by logic type, categorized by their focus on normative relationships versus descriptive conditions. (Bottom) Refinement statements by logic type, categorized by their focus on normative relationships versus descriptive conditions.}
\label{tab:logic-refinements}
\end{table*}

\paragraph{Domain-Specific Performance}

Figure~\ref{fig:domain-performance-percent-capped} reports aggregate performance across all tasks, broken down by reasoning domain and logic. A more detailed inspection reveals a clear domain-dependent pattern: first-order logic consistently performs best in commonsense reasoning tasks, while modal and non-classical logics ({\it KD}, {\it DDLE}, and {\it DDL\_CJ}) achieve superior performance in domains involving modalities, default reasoning, and (bio)ethical reasoning. This distinction is obscured when only aggregate results are considered—for instance, while {\it KD} may outperform {\it FOL} overall, this advantage does not hold uniformly across domains. Instead, the results highlight that different logics exhibit distinct strengths, and that the choice of logic is a fundamental design decision that should be guided by the targeted reasoning domain.

\subsection{Refinement Strategies Across Formalisms}

Tables~\ref{tab:logic-refinements} and report the feedback provided by the model on how to refine the explanation in the Example in ~\ref{ex:autonomy}. Each approach requires different ways to integrate the initial explanation with explicit bridging statements to complete the proof. We focus on GPT-4o model and the first refinement.
In the example, two ethical principles generate a conflict between prima-facie reasons.\footnote{The notion of ``reason'' is central to contemporary meta-ethics \cite{schroeder2024fundamentals, tucker2025weight}. The idea of prima-facie obligation can be traced back to \cite{ross2002right}.} However, one of them is undercut (i.e., it is proven to not be relevant because of some exceptional circumstance), and therefore the remaining one is binding. 




\paragraph{Deontic/Normative Logics (DDL\_CJ and KD)} focus on the reasoning step from that moves from preliminary moral considerations (the reasons) to the deontic verdict in the hypothesis. Both refinement strategies concern the resolution of conflict between reasons. {\it DDL\_CJ} introduces a \emph{preference rule}, that establish that, under certain conditions, one of the reason ``takes precedence.'' {\it KD} undertakes a different path: it establishes a logical implication between the undercut and the obligation expressed in the hypothesis. The two refinements share the emphasis on the obligations, over the descriptive features of the circumstance of choice. Also, both refinements strategies target the last reasoning step: the resolution of conflict and detachment of the all-things-considered obligation.

\paragraph{Descriptive/Factual Logics (FOL and DDLE)} 
focus on the reasoning step that identifies moral reasons by recognizing certain morally relevant facts. Therefore, they emphasize \emph{conditions and exceptions}, trying to resolve conflict by specifying that one reason does not hold under certain exceptional conditions. {\it FOL} with Neodavidsonian semantics requires explicit \emph{exception clauses} and qualifiers (``unless,'' ``especially when'') to handle defeasible reasoning within event-based representations, as demonstrated by its refinement adding explicit exception conditions to existing statements. {\it DDLE}, meanwhile, demands explicit \emph{validity conditions} that specify when reasons count as valid considerations in deliberation, requiring statements about what makes a reason ``not valid''. Both approaches reveal that what appears as a simple factual statement in natural language (``a refusal without competence doesn't count'') requires multiple layers of explicit formal encoding to function within a proof system.

\section{Discussion}
Taken together, our qualitative and quantitative results demonstrate that logical formalisms have a significant influence on both the structure and efficiency of LLM-driven neuro-symbolic reasoning. 

Addressing \textit{RQ1}, we observe that different logical formalisms tend to localize missing reasoning steps at distinct points in the explanatory chain: deontic logics such as KD and DDL\_CJ focus refinement on the normative transition from competing reasons to an all-things-considered obligation, whereas logics such as FOL and DDLE tend to emphasize additional factual conditions or validity constraints that determine whether a consideration counts as a reason at all. 
This distinction underlies \textit{RQ2}, where logic-external approaches, particularly FOL, represent moral reasons only implicitly -- as predicates or propositional constants -- thereby weakening the logical connection between reasons and obligations and necessitating fragile, ad hoc refinements. In contrast, logic-internal approaches, particularly DDL\_CJ, represent reasons directly as conditional norms, enabling conflict resolution and facilitating the detachment of unconditional obligations (see Appendix A.2 for more details). 
With respect to \textit{RQ3}, these structural differences translate into measurable gains: logics with stronger internal normative structure converge more reliably, require fewer refinement steps, and achieve higher explanation success rates for normative cases. 

More broadly, our results suggest that logic-parametric architectures do not merely improve performance metrics but reveal how different formalisms privilege distinct stages of practical reasoning, motivating future systems that dynamically select or combine distinct logical formalisms.

\label{sec: discussion}

\section{Related Work}
\paragraph{Neuro-Symbolic NLI}
Contemporary neuro-symbolic NLI systems aim to combine the language fluency and contextual awareness of large language models (LLMs) with the rigour and transparency of formal reasoning \cite{quan-etal-2025-faithful,quan-etal-2025-peirce,quan-etal-2024-verification,pan-etal-2023-logic,olausson-etal-2023-linc,ye2023satlm,ranaldi-etal-2025-improving,arakelyan-etal-2025-flare,tan-etal-2025-enhancing-logical,qi2025large}. Recent work has explored using LLMs for tasks such as autoformalization \cite{wu2022autoformalization,zhang-etal-2025-autoformalization}, and explanation generation \cite{quan-etal-2024-verification,dalal-etal-2024-inference}, often in tandem with automated theorem provers (TPs) to verify inference validity \cite{pan-etal-2023-logic,olausson-etal-2023-linc,jiang2023draft,quan-etal-2024-verification}. In this setting, LLMs generate candidate formal representations, which are then checked, refined, or completed by logic-based components such as Isabelle/HOL or Lean. While promising, most of these architectures assume a fixed logical framework --typically first-order or propositional logic -- over which reasoning is performed. This restricts the system’s adaptability to domains that require more specialized inferential structures. Our work has a similar motivation to \citet{xu2025adaptive}; however, their focus is on dynamic solver composition rather than the impact of different logical formalisms.

\paragraph{Deontic Explanations and Formal Logic}
The notion of explanation can be informally defined as the answer to a ``Why''-question. In the context of deontic logic, why-questions may concern certain deontic verdicts, e.g.: ``Why is A obligatory(/permitted/forbidden) ?.'' Explanations in deontic logic can also take a contrastive form: ``Why A is obligatory rather than B, despite a prima-facie obligation toward B ?.''  Contrastive explanations typically involve dealing with moral conflict, exceptions, preferences, and contrary-to-duties. 
From the point of view of formal logic, the aim of providing deontic explanations is to settle certain desiderata on the system. In particular, it means that you do not just want the system to give the correct outputs, but also to be able to provide transparent, precise and convincing  motivations on why such outputs obtain. Most formal work on deontic explanation is related to the field of formal argumentation \cite{governatori2022stable, rotolo2023argumentation, van2024towards}. 

\section{Conclusion}
 

This paper introduced a logic-parametric framework for neuro-symbolic NLI, leveraging the LogiKEy methodology to embed diverse logical systems within higher-order logic. Our central claim is that the choice of logic is not a neutral design decision, but a critical factor influencing inference generalizability, proof behavior, and explanation quality.



The broader implication of this work is the possibility of building \textit{logic-adaptive reasoning architectures} -- systems capable of selecting, switching, or combining logics dynamically based on the needs of a given task or input. Instead of assuming a fixed logic for all reasoning processes, such pipelines would treat logic selection analogously to model selection in machine learning, guided by structural, semantic, or contextual cues.

\section{Limitations}

While our analysis provides insights into how different logical formalisms interacting LLMs handle normative and ethical reasoning, several limitations must be acknowledged.

Our study utilizes  two language models (GPT-4o and DeepSeek-V1) for generating and refining logical formalizations. Further work is required to understand how our findings generalize to different models. For example, smaller models might struggle with the complex multi-step reasoning required, while other large models might employ different inference strategies.

In addition, our empirical investigation examines four specific logical systems ({\it KD}, {\it FOL}, {\it DDLE}, and {\it DDL\_CJ}). These represent particular approaches to normative reasoning, but numerous alternative formalisms exist --including quantified modal and conditional logics, default logics, argumentation frameworks, non-monotonic logics, and probabilistic reasoning systems. Future work can investigate how our observations apply to other systems, each of which might require different types of refinements or exhibit different failure modes. 

Furthermore, the refinement strategies analyzed were generated by automated systems or through structured interactive processes. These refinements represent \textit{one possible path} to completing each proof, but alternative refinements might also resolve the logical gaps. We have not systematically explored the space of all possible refinements for each formalism, nor have we evaluated whether the chosen refinements represent the most natural solutions from a human reasoning perspective.

Finally, our study uses relatively simple, structured natural language explanations to enable full experimental control. Real-world reasoning might involve more complex, nuanced, or implicit reasoning patterns that may present additional challenges for formalization. The limitations we observe might be amplified with more complex natural language inputs, particularly those involving ambiguous terms, contextual dependencies, or culturally specific normative concepts.

\bibliography{custom}

\appendix

\section{Appendix}
\label{sec:appendix_prompts}
\subsection{Prompt Adaptation for Logic-Parametric Formalization}
\label{sec:prompt-adaptation}

To support logic-parametric theorem proving, we adapt the prompting framework from \citet{quan-etal-2025-faithful}, originally designed for Neo-Davidsonian first-order logic (FOL) formalization. While their prompts effectively guide LLMs toward faithful event-based representations, they assume a fixed logical substrate. Our extension modifies these prompts to accommodate a range of classical and non-classical logics embedded via the LogiKEy framework. 

\begin{figure*}[h!]
 
    \centering

    \begin{subfigure}{0.95\textwidth}
        \centering
        \includegraphics[width=\textwidth]{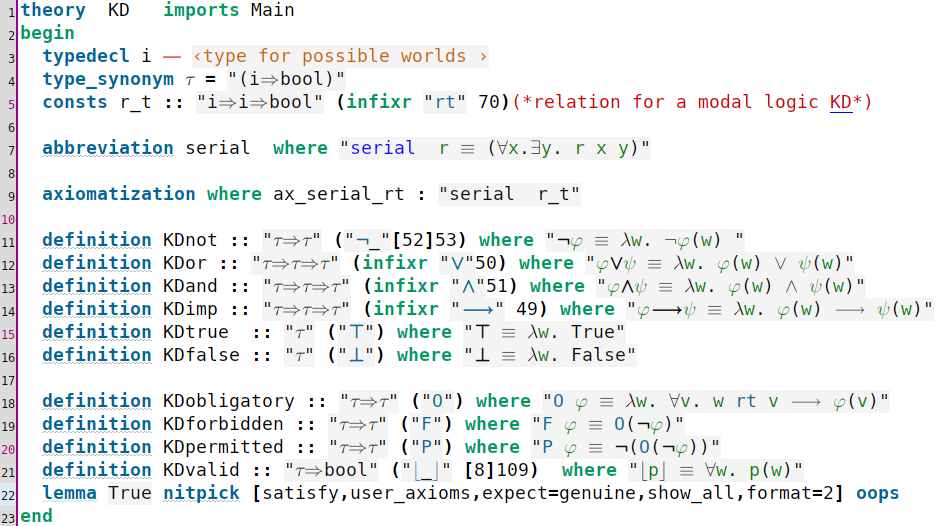}
        \caption{Modal Logic KD in HOL.}
        \label{fig:kd-hol}
    \end{subfigure}

    \vspace{-0.6em}
    {\small
    \begin{itemize}[leftmargin=*, itemsep=0pt, topsep=2pt, parsep=0pt]
        \item Line 3 introduces the primitive type $i$ for possible worlds.
        \item Line 4 introduces the type $\tau$ for formulas.
        \item Line 5 introduces $r_t$, encoding the accessibility relation.
        \item Line 7 restricts the accessibility relation by seriality.
        \item Lines 11--16 define Boolean connectives.
        \item Lines 18--20 define the monadic deontic operators (obligation, forbidden, permission).
        \item Line 21 introduces global validity.
        \item Line 22 uses Nitpick to confirm consistency.
    \end{itemize}
    }


    \begin{subfigure}{0.49\textwidth}
        \centering
        \includegraphics[width=\textwidth]{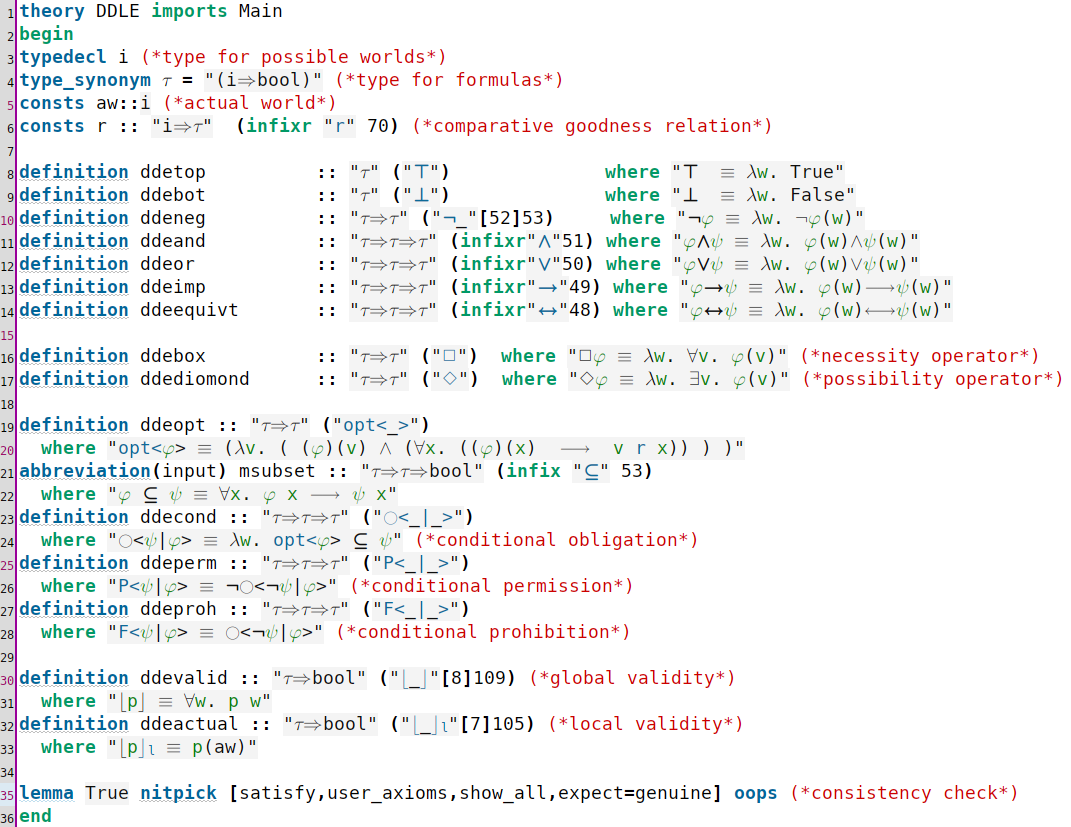}
        \caption{Conditional Logic DDLE in HOL.}
        \label{fig:ddle-hol}
    \end{subfigure}
    \hfill
    \begin{subfigure}{0.49\textwidth}
        \centering
        \includegraphics[width=\textwidth]{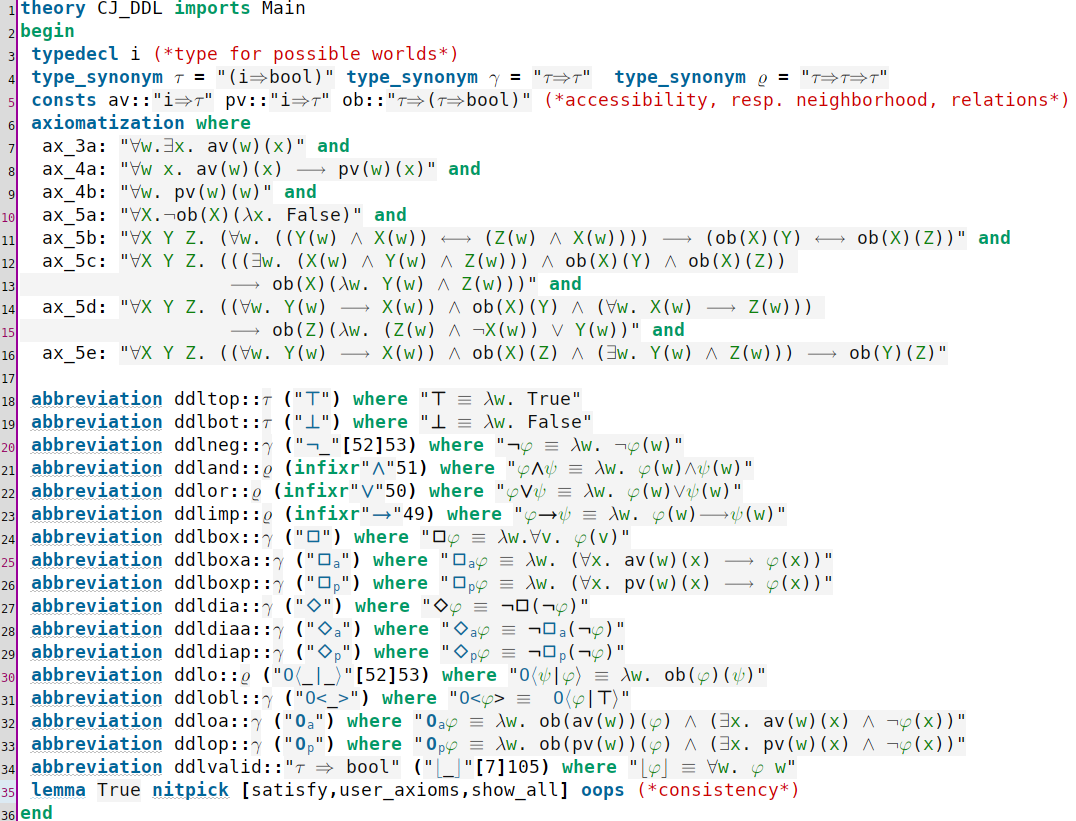}
        \caption{Conditional Logic DDL\_CJ in HOL.}
        \label{fig:ddlcj-hol}
    \end{subfigure}

    \vspace{-0.3em}
    \caption{Isabelle/HOL embeddings used in our logic-parametric setting: KD (top) and two conditional logics (bottom).}
    \label{fig:hol-embeddings}
\end{figure*}

 \paragraph{Modal and Conditional Logic in HOL} The so-called \textit{shallow semantical embedding} approach was developed by Benzmüller \cite{J41} for translating the semantics of classical and non-classical logics into HOL. The embedding of a broad class of modal logics in HOL is discussed in \cite{J21,J23}. The semantic embedding of dyadic deontic logic (system {\bf DDLE}) is covered in \cite{J45}, and the embedding of Carmo and Jones conditional logic is presented in \cite{C71}. Figure~\ref{fig:hol-embeddings} discusses these logical embeddings in HOL in more detail.

\paragraph{Overview of Prompt Adaptation.}
Rather than introducing entirely new prompts, we retain the overall structure of the original pipeline and selectively adapt prompts whose behaviour depends on the underlying logic. This allows us to preserve logic-invariant semantic extraction while enabling logic-specific reasoning, refinement, and proof construction. Concretely, we distinguish between prompts that are \emph{logic-agnostic} and those that are \emph{logic-sensitive}, and only the latter are modified when changing the target logic.

\paragraph{Logic-Agnostic Prompts}
The first class of prompts is responsible for extracting semantic content from natural language, independent of the target logic. These prompts include:
\begin{itemize}
    \item \textbf{Syntactic Parsing Prompts} -- Extract grammatical structure to guide predicate-argument mapping.
    \item \textbf{Generate  and Refine Explanation prompts}: Asking LLMs to use causal knowledge and commonsense to provide logical explanations for the provided causal reasoning scenarios.
\end{itemize}

Since these prompts operate purely form natural language inference and at the level of semantic parsing, their output remains stable across different logical settings. As a result, they are reused unchanged across all experiments. This design choice ensures that differences observed across logics are attributable to reasoning and proof mechanisms rather than to variations in semantic interpretation.

\paragraph{Logic-Sensitive Encoding Prompts}
The second class of prompts translates semantic representations into Isabelle/HOL axioms and theorem statements. These prompts are directly affected by the choice of logic, as they determine the logical operators, modal or deontic constructs, and axiom schemas used in the formalization.

    
    
    

\begin{figure*}[h!]
    \centering
    \begin{subfigure}{0.92\textwidth}
        \centering
        \includegraphics[width=0.95\textwidth]{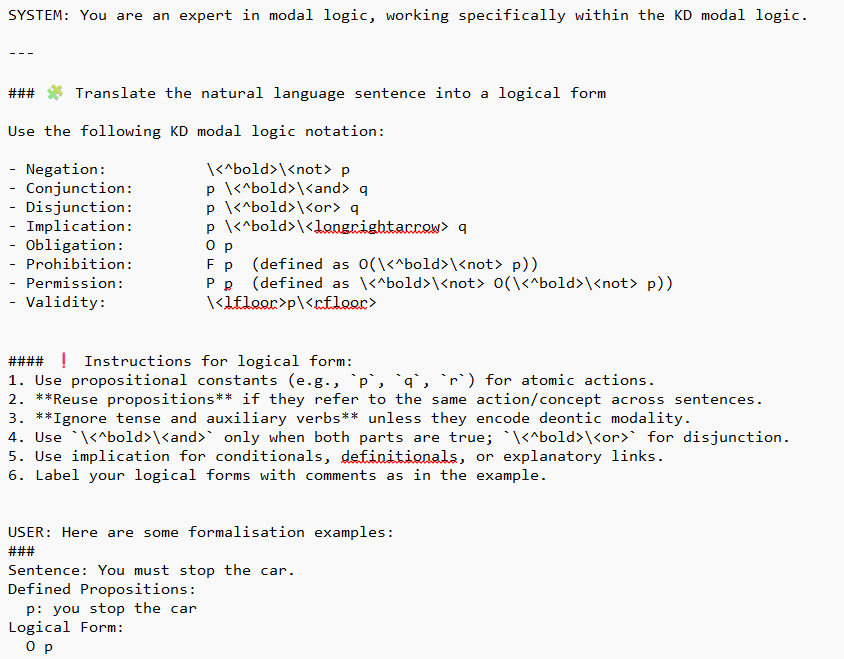}
        \caption{KD-Syntax prompt.}
        \label{fig:kd-syntax-prompt}
    \end{subfigure}


    \begin{subfigure}{0.92\textwidth}
        \centering
        \includegraphics[width=0.95\textwidth]{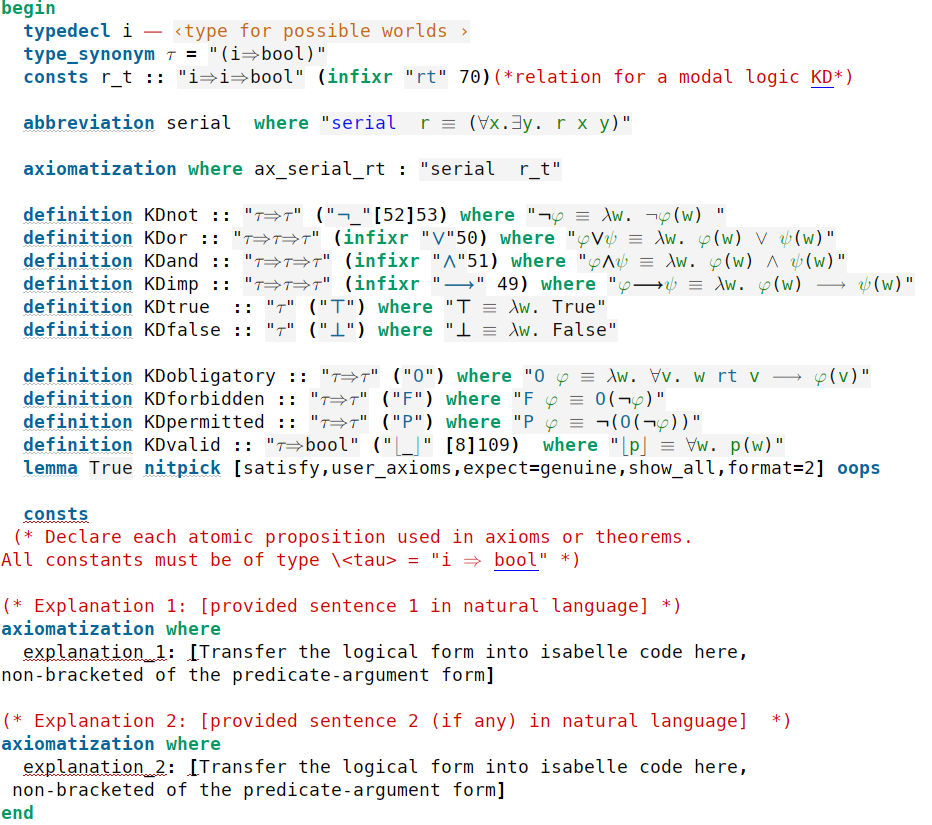}
        \caption{KD Isabelle axiom prompt (includes the KD-in-HOL theory snippet).}
        \label{fig:kd-isabelle-axiom-prompt}
    \end{subfigure}

    \vspace{-0.3em}
    \caption{Prompts used for KD: the syntax-check prompt (top) and the Isabelle-axiom prompt (bottom), which embeds the KD theory in HOL.}
    \label{fig:kd-prompts}
\end{figure*}

Importantly, while the syntactic form of the generated axioms changes with the logic, the underlying semantic content extracted from the input remains fixed. This separation allows us to systematically study the effect of logic choice on downstream reasoning. As illustrated in Figure~\ref{fig:kd-prompts}, the auto-formalization of natural language into modal logic relies on the encoding of {\bf KD} in HOL, and the Isabelle-axiom prompt is grounded in the same underlying {\bf KD} theory.

\paragraph{Logic-Sensitive Refinement and Repair Prompts}
A third class of prompts handles explanation refinement, contradiction resolution, and syntactic repair when proof attempts fail. These prompts are highly logic-dependent, as the notion of contradiction and the admissible repair strategies vary substantially across logics. These adaptations play a central role in the observed differences in refinement efficiency and robustness across logics, with modal and conditional logics requiring more nuanced error correction strategies.

    
    
    

\paragraph{Logic-Sensitive Proof Construction Prompts}
Finally, proof construction prompts generate Isabelle/HOL proof sketches based on the axioms and hypotheses produced earlier in the pipeline. These prompts are the most sensitive to logic choice, as proof strategies depend directly on the inference rules and axioms of the target logic. 
For each logics, we modify these prompts to ensure that generated proofs respect the corresponding modal or deontic principles encoded in LogiKEy.


\subsection{Reasoning with Moral Reasons: A Comparison of the Four Logics}
\paragraph{}In the explanation of Example~\ref{ex:autonomy}, the core inference pattern is an undercut. It is composed of two steps: first, two prima facie reasons are instantiated. Second, one of them is undercut. The premises that introduce the reasons exhibit a typical scheme in bioethical reasoning: “Given Principle1, Fact1($F1$) is a reason for Act1($A1$)”. It’s interesting to see how the four logics formalize this proposition. Note that none of them has a primitive operator to express reasons. 
\begin{itemize}
    \item In {\bf FOL}, “...is a reason for…” is treated as a predicate. 
\item In {\bf KD}, the deontic operator is used to express the general command to respect autonomy and the deontic verdict in the hypothesis, but not the reason-relation between $F1$ and $A1$. ``$F1$ is a reason for $A1$'' is captured using a propositional constant.
\item In {\bf DDLE}, in the first iteration, the dyadic deontic operator is used to express the reason as a conditional norm: $\bigcirc(A1 / F1)$. However, the refinement procedure induces a flat representation of the reason as, again, a propositional constant. 
\end{itemize}
 When reasons are represented so abstractly, their relation with the all-things-considered obligation is lost. The model needs to capture such relation by introducing new explanatory sentences that establish the logical connection, but this process can be long and unreliable. In fact, none of the three logics mentioned so far is able to provide a valid explanation in the case at stake.  

\begin{itemize}
    \item {\bf DDL\_CJ}, in contrast, manages to use the dyadic deontic operator to capture reasons. Both the reason for $A1$ and the reason against are captured as conditional norms. 
\end{itemize}
This enables to represent reasons, but how to get the all-things-considered obligation? In the present example, the key is the undercut, that works as a resolution technique to dissolve the conflict between the two reasons. This reasoning step is typical of defeasible logics and argumentation, but not exactly what {\bf DDL\_CJ} is designed for. However, the hybrid model manages to overcome the obstacle, rephrasing the explanation as a conditional preference: “When $F2$ is the case, the reason for $A1$ takes precedence”.  This proposition is then captured as, in turn, a conditional obligation: $\bigcirc(A1 / F2)$; i.e., ``Given that $F2$, it ought to be that $A1$.'' This strategy pays off: {\bf DDL\_CJ} is able to represent both reasons as conditional norms, and to detach an unconditional obligation to give the treatment, thus reaching a valid explanation. 
\paragraph{}This analysis shows that the LLM – at least in some cases – is able to push the logic beyond its preferred interpretative domain. The deontic operator can be used to reconstruct normative notions such as the moral reasons. When this happens, the otherwise tedious task to handcraft the relations between different normative notions (e.g. the connection between what you have most reason to do and what you should do) is partially outsourced to the theorems of the logic, increasing the chances of reaching a valid explanation quickly.

\subsection{More Details on the Dataset} 
\paragraph{}As we explain in Section 3, the dataset explores deontic and – more specifically – ethical reasoning. Our interest lies in the different reasoning patterns that compose it. The organization of the dataset reflects this focus: each folder targets a specific inference pattern or combination of patterns. In this subsection, we present all folders. For each of them, we specify and comment on the corresponding inference pattern. 
\paragraph{} Before presenting the folders, it is useful to recall the structure of the cases. Each case is composed of a set of premises, an hypothesis and an explanation. The role of the explanation is to bridge the premises and the hypothesis through a series of reasoning steps. These steps determine to which folder the case belongs. 
\begin{enumerate}
    \item \textbf{Classical logic (5 cases).}\footnote{In this folder, we adapt 3 cases from \cite{holliday2024conditional}.} The cases in this folder are such that applying classical logic on the premises is sufficient to infer the hypothesis. 
    \item \textbf{Common sense (10 cases).}\footnote{In this folder, we adapt 5 cases from the e-SNLI dataset \cite{camburu2018snli}.} This folder is divided into two sub-folders: the former explores epistemic common sense reasoning, while the latter focuses on practical common sense reasoning. In both cases, the key to infer the hypothesis is to understand that two terms have equivalent meanings. Note that, in natural language, this can happen with deontic statements, too: one can, for instance, say that ``Killing is wrong'', or that ``One shall not kill''.\footnote{These different expressions are studied and grouped into families by meta-ethicists \cite{berker2022deontic}. In this folder, we only use couples of terms that belong to the family of deontic categories. In the next one on default reasoning, we use couples that range across different families.}
    \item \textbf{Default reasoning (17 cases).}\footnote{In this folder, we adapt 5 cases from the e-SNLI dataset. } This folder is divided in two sub-folders: ``Epistemic default reasoning'' and ``Practical default reasoning''. The former contains cases in which some default assumption about the world must be used to infer the hypothesis. In the latter, we put two types of cases. First, we put the practical counterpart of the epistemic cases, that is: cases in which one must use a default assumption about morality (e.g., that one should not do what is bad), to infer the hypothesis. Second, we introduce the notion of a moral reason. A reason is a fact that speaks in favor or against a certain action \cite{schroeder2024fundamentals,tucker2025weight}. In the cases of this second type, to explain the deontic verdict in the hypothesis one must recognize that some fact in the premises is a reason that support it. 
    \item \textbf{Modalities (24 cases).}\footnote{In this folder, we adapt 2 cases from \cite{holliday2024conditional}.} The alethic modalities (necessity, possibility, impossibility) and the deontic modalities (obligation, permission, prohibition) display certain logical relations: for instance, if $\varphi$ is necessary then $\neg \varphi$ is not possible, and similarly, if $\varphi$ is obligatory then $\neg \varphi$ is not permissible. In the cases of this folder, the explanation refers to the logical relations between modalities. Since our focus is on practical reasoning, we devote more space to deontic modalities. In particular, we introduce the notion of conditional obligation, central in dyadic deontic logics and usually expressed formally with the notation: $\bigcirc(\psi / \varphi)$. We explore two reasoning patterns related to conditional oughts: factual detachment (i.e., from $\bigcirc(\psi / \varphi)$ and $\varphi$, $\bigcirc(\psi / \top)$ is inferred), and deontic detachment (i.e., from $\bigcirc(\psi / \varphi)$ and $\bigcirc(\varphi / \top)$, $\bigcirc(\psi / \top)$ is inferred).
    \item \textbf{(Bio)ethics.} Suppose you endorse some general moral principles, such as that benevolence is good and autonomy should be respected. Now consider a bioethical case in which a patient refuses a beneficial treatment. Should you force the treatment? To answer, you need to use the general ethical principles to identify prima facie reasons (step 1). For example, given that autonomy should be respected, the patient's refusal is a reason not to force the treatment. Then, if you end up with conflicting reasons, you must find a way to solve the conflict\footnote{A remark is in order here: We are not making any moral claim about how the cases should be evaluated. Rather, we focus on what explains the decision, whatever the decision is.} and infer an all-things-considered obligation (step 2).\footnote{Although inspired by principlism \cite{beauchamp2019principles}, this framework is very general: it can express any theory of ethical reasoning that presents some general principles, rules, or duties that are used to identify prima facie obligations or reasons in specific circumstances of choice.} Given this characterization of ethical reasoning, we organize the folder in order to explore the different reasoning patterns at play: 
    
    \begin{enumerate}
        \item \textbf{From principles to prima facie reasons (12 cases).} The cases in this sub-folder are such that there is only one relevant reason. This makes step 2 of ethical reasoning trivial and allows us to focus on step 1.
        \item \textbf{Undercuts (3 cases).} Recall the example of the patient who refuses the treatment, but now suppose that the patient is not competent because of high fever. In this case, you should not consider the patient's refusal as a valid reason. Using the terminology of the argumentation community \cite{baroni2018handbook}, we say that the reason is \emph{undercut}. The cases in this folder focus on undercuts.
        \item \textbf{Conflict within one principle (9 cases).} Imagine you think that the only moral principle is benevolence. You can still have conflict of reasons: sometimes you cannot be benevolent to everybody even though, ideally, you should (think of the ethical issues around resource allocation). The cases in this folder explore such conflicts, varying the resolution techniques (weighing of reasons, case-based reasoning, undercuts).
        \item \textbf{Conflict across different principles (10 cases).} The example of the patient who refuses treatment is a case of conflict between reasons that refer to different principles. In this sub-folder, we explore these conflicts, varying the resolution techniques.
        \item \textbf{Case-study: euthanasia (13 cases).} The cases in this sub-folder form a roster of possible scenarios concerning the ethical issues around the practice of euthanasia. Their contribution to the repository consists in their complexity: they aim to approximate the richness and ambiguity of real-life choices. 
    \end{enumerate}
\end{enumerate}


\FloatBarrier
 
\begin{table*}[t]
\centering
\small
\setlength{\tabcolsep}{3pt}
\renewcommand{\arraystretch}{0.90}
\begin{tabular}{lcc}
\hline
\textbf{Logic} & \textbf{DeepSeek (\%)} & \textbf{ChatGPT (\%)} \\
\hline
FOL      & 42.72 & 39.81 \\
KD       & 76.70 & 77.67 \\
DDLE     & 6019.0 & 51.46\\
DDL\_CJ  & 61.17 & 74.76 \\
\hline
\end{tabular}
\caption{Success rates for valid explanation generation. }
\label{tab:success-rates}
\end{table*}

\begin{table*}[t]
\centering
\small
\setlength{\tabcolsep}{3pt}
\renewcommand{\arraystretch}{0.90}
\begin{tabular}{lcc}
\hline
\textbf{Logic} & \textbf{DeepSeek} & \textbf{ChatGPT} \\
\hline
FOL      & 0.91 & 0.83 \\
KD       & 0.43 & 0.60 \\
DDLE     & 0.60 & 0.64 \\
DDL\_CJ  & 0.65 & 0.71 \\
\hline
\end{tabular}
\caption{Average number of refinement iterations required to reach a valid explanation.
}
\label{tab:iterations}
\end{table*}

\begin{table*}[t]
\centering
\small
\setlength{\tabcolsep}{3pt}
\renewcommand{\arraystretch}{0.90}
\begin{tabular}{lcc}
\hline
\textbf{Logic} & \textbf{DeepSeek (s)} & \textbf{ChatGPT (s)} \\
\hline
FOL      & 108.99 & 103.74 \\
KD       & 65.25  & 58.83  \\
DDLE     & 69.45  & 65.34  \\
DDL\_CJ  & 60.18  & 49.80  \\
\hline
\end{tabular}
\caption{Average solving time (seconds) over successful explanation refinements.
}
\label{tab:runtime}
\end{table*}
 
\begin{table*}[t]
\centering
\small
\setlength{\tabcolsep}{3pt}
\renewcommand{\arraystretch}{0.90}
\begin{tabular}{lcc}
\hline
\textbf{Logic} & \textbf{DeepSeek (\%)} & \textbf{ChatGPT (\%)} \\
\hline
FOL      & 3.90 & 2.90 \\
KD       & 7.80 & 1.90  \\
DDLE     & 16.50 & 12.60 \\
DDL\_CJ  & 19.40  & 5.80  \\
\hline
\end{tabular}
\caption{Syntactic error rates. ChatGPT-KD has lowest error rate (less than 2\%), demonstrating superior robustness.}
\label{tab:syntactic-errors}
\end{table*}

\begin{table*}[t]
\centering
\scriptsize
\setlength{\tabcolsep}{3pt}
\renewcommand{\arraystretch}{0.92}
\begin{tabular}{lccccc}
\hline
\textbf{Domain} & \textbf{Logic} & \textbf{Model} & \textbf{Valid (\%)} & \textbf{Avg. Itr} & \textbf{Time (s)} \\
\hline
\multirow{1}{1cm}{\textbf{Classical Logic (5)}} 
 & FOL & DeepSeek & 100.0 & 0.20 & 64.83 \\
 & FOL & ChatGPT & 80.0 & 0.25 & 46.54 \\
 & KD & DeepSeek & 100.0 & 0.00 & 22.09 \\
 & KD & ChatGPT & 80.0 & 0.75 & 41.64 \\
 & DDLE & DeepSeek & 100.0 & 0.00 & 24.65 \\
 & DDLE & ChatGPT & 80.0 & 0.75 & 52.49 \\
 & DDL\_CJ & DeepSeek & 80.0 & 0.50 & 11.03 \\
 & DDL\_CJ & ChatGPT & 60.0 & 0.67 & 26.11 \\
\hline
\multirow{1}{1cm}{\textbf{Common Sense (10)}}
 & FOL & DeepSeek & 90.0 & 1.33 & 83.53 \\
 & FOL & ChatGPT & 80.0 & 0.75 & 55.60 \\
 & KD & DeepSeek & 70.0 & 0.57 & 66.44 \\
 & KD & ChatGPT & 50.0 & 1.00 & 81.30 \\
 & DDLE & DeepSeek & 70.0 & 0.28 & 47.36 \\
 & DDLE & ChatGPT & 60.0 & 0.50 & 65.49 \\
 & DDL\_CJ & DeepSeek & 40.0 & 0.85 & 59.33 \\
 & DDL\_CJ & ChatGPT & 50.0 & 0.80 & 63.05 \\
\hline
\multirow{1}{1cm}{\textbf{Default Reasoning (17)}}
 & FOL & DeepSeek & 52.94 & 1.11 & 105.21 \\
 & FOL & ChatGPT & 70.59 & 1.17 & 122.81 \\
 & KD & DeepSeek & 82.35 & 0.50 & 58.39 \\
 & KD & ChatGPT & 76.74 & 0.62 & 65.96 \\
 & DDLE & DeepSeek & 58.82 & 0.40 & 68.61 \\
 & DDLE & ChatGPT & 64.71 & 0.55 & 52.51 \\
 & DDL\_CJ & DeepSeek & 52.94 & 1.11 & 78.73 \\
 & DDL\_CJ & ChatGPT & 88.24 & 0.93 & 46.49 \\
\hline
\multirow{1}{1cm}{\textbf{Modalities (24)}}
 & FOL & DeepSeek & 58.33 & 0.86 & 89.41 \\
 & FOL & ChatGPT & 45.83 & 0.54 & 70.10 \\
 & KD & DeepSeek & 87.50 & 0.00 & 47.45 \\
 & KD & ChatGPT & 91.67 & 0.32 & 53.01 \\
 & DDLE & DeepSeek & 62.50 & 0.73 & 36.56 \\
 & DDLE & ChatGPT & 58.33 & 0.71 & 55.45 \\
 & DDL\_CJ & DeepSeek & 62.50 & 0.53 & 38.36 \\
 & DDL\_CJ & ChatGPT & 62.50 & 0.40 & 43.46 \\
\hline
\multirow{1}{1cm}{\textbf{(Bio)ethics (47)}}
 & FOL & DeepSeek & 14.89 & 0.71 & 130.01 \\
 & FOL & ChatGPT & 12.77 & 1.17 & 129.46 \\
 & KD & DeepSeek & 68.09 & 0.72 & 82.62 \\
 & KD & ChatGPT & 76.60 & 0.69 & 56.91 \\
 & DDLE & DeepSeek & 53.19 & 0.80 & 94.05 \\
 & DDLE & ChatGPT & 38.30 & 0.67 & 76.75 \\
 & DDL\_CJ & DeepSeek & 65.96& 0.61 & 69.05 \\
 & DDL\_CJ & ChatGPT & 82.98 & 0.74 & 53.17 \\
\hline
\end{tabular}
\caption{Aggregate explanation refinement performance across logical frameworks for both DeepSeek and ChatGPT models. Solving time is averaged over successful runs only.}
\label{tab:domain-performance-percent-capped}
\end{table*}

\label{sec:appendix_tables}
\FloatBarrier

\end{document}